\documentclass[10pt,twocolumn]{article}

\usepackage{graphicx}
\usepackage{algorithm,algorithmic}
\usepackage{amsmath}
\usepackage{multirow}
\usepackage{float}
\usepackage{url}
\usepackage[none]{hyphenat}
\usepackage{authblk}

\usepackage{float}
\newcommand\T{\rule{0pt}{2.6ex}}
\newcommand\B{\rule[-1.2ex]{0pt}{0pt}}
\newcommand\TT{\rule{0pt}{3.0ex}}
\newcommand\BB{\rule[-1.5ex]{0pt}{0pt}}

\begin{document}
\newcounter{cntr1}
\newcounter{cntr2}
%\brokenpenalty=10000\relax
\emergencystretch 3em

\title{Water Distribution System Design Using Multi-Objective\\
Genetic Algorithm with External Archive and Local Search}

\author[1]{Mahesh B. Patil}
\author[2]{M. Naveen Naidu}
\author[2]{A. Vasan}
\author[2]{Murari R. R. Varma}
\affil[1]{Department of Electrical Engineering, Indian Institute of Technology Bombay}
\affil[2]{Department of Civil Engineering, BITS Pilani Hyderabad Campus}

\maketitle

\begin{abstract}
Hybridisation of the multi-objective optimisation algorithm NSGA-II
and local search is proposed for water distribution system design.
Results obtained with the proposed algorithm are
presented for four medium-size water networks taken from the
literature. Local search is found to be beneficial for one of the
networks in terms of finding new solutions not reported earlier. It is also shown
that simply using an external archive to save all non-dominated
solutions visited by the population, even without local search,
leads to substantial improvement in the non-dominated set produced
by the algorithm.
\end{abstract}

\section{Introduction}
Optimisation of water distribution systems (WDS) for the dual
objectives of minimising cost and maximising network resilience
is a challenging problem because of the large
solution spaces involved (see \cite{wang2014}, \cite{moosavian2016} and
references therein). In this context, the benchmark water network
problems made available by
Wang {\it{et al.}}\,\cite{wang2014}
have served as an excellent resource for researchers trying out
new optimisation algorithms. Recently, hybridisation of local search
and the multi-objective particle swarm optimisation algorithm
(MOPSO)\,\cite{coello2004} was shown to be very effective\,\cite{patil2019}
for the two-objective WDS design problem.

Table~\ref{tbl_mopso_mbp} presents a summary of the performance
of this new ``MOPSO+"
algorithm\,\cite{patil2019} for the four medium-size water networks
given in
\cite{wang2014}. The table compares the sets of
non-dominated (ND) solutions (loosely called ``Pareto fronts" or PFs)
by two algorithms. Algorithm\,1 (called ``UExeter") is a
combination of five multi-objective
evolutionary algorithms (MOEAs) presented in
\cite{wang2014}, whereas
Algorithm\,2 is the MOPSO+ algorithm of
\cite{patil2019}.
$N_1^t$ is the total number of ND solutions obtained by algorithm 1 of
which $N_1^a$ are accepted and $N_1^r$ are rejected (since they got
dominated by some of the ND solutions given by algorithm 2). The number
of unique solutions given by algorithm 1, i.e., solutions which could
not be obtained by algorithm 2, is denoted by $N_1^u$, and the number
of common solutions between the two algorithms by $N^c$.
The total number of function evaluations over all independent runs of the
concerned algorithm is denoted by
$N_{FE}^{\mathrm{net}}$.
As seen from the table, $N_1^u$ is nearly zero in all cases which means
that the MOPSO+ algorithm has covered all solutions given by algorithm 1.
Furthermore, $N_2^u$ is significantly large, which means that algorithm 2
has produced many solutions which were not present in the ND set obtained
by algorithm 1. Comparing the
$N_{FE}^{\mathrm{net}}$
values, we see that the computational efforts for
the two algorithms are similar. In summary, the MOPSO+ algorithm has
performed better without requiring a significantly larger computational
effort.

The above beneficial hybridisation of local search
with the MOPSO algorithm opens up the
interesting possibility of improving the performance of other MOEAs
using local search. It is the purpose of this paper to explore the
effectiveness of local search when hybridised with another commonly
used MOEA, viz., the NSGA-II algorithm\,\cite{deb2002}, for the
WDS design problem described in
\cite{wang2014}.
The paper is organised as follows. In
Sec.~\ref{sec_nsga2_ls},
we describe the modifications of the basic NSGA-II algorithm to
combine it with local search. In
Sec.~\ref{sec_results},
we present results obtained with the different schemes of
Sec.~\ref{sec_nsga2_ls} for the four medium-size water networks
described in
\cite{wang2014}. Finally, we present the conclusions of this study in
Sec.~\ref{sec_conclusions}.

\begin{table*}[h]
   \centering
   \caption{Comparison of UExeter\,\cite{wang2014} and
    MOPSO+\,\cite{patil2019} non-dominated solution sets (``PFs")
    for four medium-size water networks.}

    \hspace*{0.5cm}

    \begin{tabular}{|c|r|r|r|r|r|r|r|r|r|r|r|}
    \hline
      {\multirow{2}{*}{Network}}
    & \multicolumn{5}{c|}{\B \T UExeter (PF-1)}
    & \multicolumn{5}{c|}{MOPSO+ (PF-2)}
    & {\multirow{2}{*}{$N^c$}}
    \\ \cline{2-11}
       {}
    &  \multicolumn{1}{c|}{\B\T $N_1^t$}
    &  \multicolumn{1}{c|}{\B\T $N_1^a$}
    &  \multicolumn{1}{c|}{\B\T $N_1^r$}
    &  \multicolumn{1}{c|}{\B\T $N_1^u$}
    &  \multicolumn{1}{c|}{\B\T $N_{FE}^{\mathrm{net}}$}
    &  \multicolumn{1}{c|}{\B\T $N_2^t$}
    &  \multicolumn{1}{c|}{\B\T $N_2^a$}
    &  \multicolumn{1}{c|}{\B\T $N_2^r$}
    &  \multicolumn{1}{c|}{\B\T $N_2^u$}
    &  \multicolumn{1}{c|}{\B\T $N_{FE}^{\mathrm{net}}$}
    & {}
    \\ \hline
    \B \T HAN & 575 & 534 & 41 & 1~ & 90\,M &  750 &  748 & 2~ & 215 &  74.6\,M & 533
    \\ \hline
    \B \T BLA & 901 & 849 & 52 & 0~ & 90\,M & 1045 & 1045 & 0~ & 196 &  44.1\,M & 849
    \\ \hline
    \B \T NYT & 627 & 595 & 32 & 4~ & 90\,M &  661 &  656 &  5~ & 65 & 130.3\,M & 591
    \\ \hline
    \B \T GOY & 489 & 444 & 45 & 3~ & 90\,M &  571 &  570 & 1~ & 129 &  37.9\,M & 441
    \\ \hline
    \end{tabular}
\label{tbl_mopso_mbp}
\end{table*}

\begin{table*}[h]
   \centering
   \caption{Comparison of PFs obtained in \cite{wang2014} and algorithm D
    for different values of $N_{\mathrm{link}}$.}

    \hspace*{0.5cm}

    \begin{tabular}{|c|c||r|r|r|r||r|r|r|r||r|}
    \hline
      {\multirow{2}{*}{Network}}
    & {\multirow{2}{*}{$N_{\mathrm{link}}$}}
    & \multicolumn{4}{c||}{\B \T UExeter (PF-1)}
    & \multicolumn{4}{c||}{Scheme D (PF-2)}
    & {\multirow{2}{*}{$N^c$}}
    \\ \cline{3-10}
      {}
    & {}
    &  \multicolumn{1}{c|}{\B\T $N_1^t$}
    &  \multicolumn{1}{c|}{\B\T $N_1^a$}
    &  \multicolumn{1}{c|}{\B\T $N_1^r$}
    &  \multicolumn{1}{c||}{\B\T $N_1^u$}
    &  \multicolumn{1}{c|}{\B\T $N_2^t$}
    &  \multicolumn{1}{c|}{\B\T $N_2^a$}
    &  \multicolumn{1}{c|}{\B\T $N_2^r$}
    &  \multicolumn{1}{c||}{\B\T $N_2^u$}
    & {}
    \\ \hline
{\multirow{4}{*}{{\rule{0pt}{7.0ex}} HAN}}
& \BB \TT 1
& 575 & 547 & 28 & 44
& 692 & 659 & 33 & 156
& 503
\\ \cline{2-11}
{} 
& \BB \TT 10
& 575 & 547 & 28 & 4
& 707 & 702 & 5 & 159
& 543
\\ \cline{2-11}
{} 
& \BB \TT 50
& 575 & 545 & 30 & 4
& 713 & 706 & 7 & 165
& 541
\\ \cline{2-11}
{} 
& \BB \TT 100
& 575 & 538 & 37 & 3
& 713 & 708 & 5 & 173
& 535
\\ \hline
{\multirow{4}{*}{{\rule{0pt}{7.0ex}} BLA}}
& \BB \TT 1
& 901 & 851 & 50 & 33
& 1023 & 1000 & 23 & 182
& 818
\\ \cline{2-11}
{} 
& \BB \TT 10
& 901 & 849 & 52 & 0
& 1040 & 1040 & 0 & 191
& 849
\\ \cline{2-11}
{} 
& \BB \TT 50
& 901 & 849 & 52 & 0
& 1036 & 1036 & 0 & 187
& 849
\\ \cline{2-11}
{} 
& \BB \TT 100
& 901 & 849 & 52 & 0
& 1040 & 1040 & 0 & 191
& 849
\\ \hline
{\multirow{4}{*}{{\rule{0pt}{7.0ex}} NYT}}
& \BB \TT 1
& 627 & 591 & 36 & 30
& 643 & 631 & 12 & 70
& 561
\\ \cline{2-11} 
{}
& \BB \TT 10
& 627 & 591 & 36 & 22
& 648 & 640 & 8 & 71
& 569
\\ \cline{2-11} 
{}
& \BB \TT 50
& 627 & 591 & 36 & 22
& 647 & 640 & 7 & 71
& 569
\\ \cline{2-11} 
{}
& \BB \TT 100
& 627 & 591 & 36 & 22
& 648 & 640 & 8 & 71
& 569
\\ \hline
{\multirow{4}{*}{{\rule{0pt}{7.0ex}} GOY}}
& \BB \TT 1
& 489 & 448 & 41 & 89
& 521 & 465 & 56 & 106
& 359
\\ \cline{2-11}
{} 
& \BB \TT 10
& 489 & 444 & 45 & 56
& 535 & 510 & 25 & 122
& 388
\\ \cline{2-11}
{} 
& \BB \TT 50
& 489 & 444 & 45 & 55
& 526 & 510 & 16 & 121
& 389
\\ \cline{2-11}
{} 
& \BB \TT 100
& 489 & 444 & 45 & 56
& 544 & 509 & 35 & 121
& 388
\\ \hline
\end{tabular}
    \hspace*{0cm}
\label{tbl_nlink}
\end{table*}

\begin{table*}[h]
   \centering
   \caption{Comparison of PFs obtained in \cite{wang2014} and algorithms A-D.}

    \hspace*{0.5cm}

    \begin{tabular}{|c|c||r|r|r|r|r||r|r|r|r|r||r|}
    \hline
      {\multirow{2}{*}{Network}}
    & {\multirow{2}{*}{Algorithm}}
    & \multicolumn{5}{c||}{\B \T UExeter (PF-1)}
    & \multicolumn{5}{c||}{Algorithm A/B/C/D (PF-2)}
    & {\multirow{2}{*}{$N^c$}}
    \\ \cline{3-12}
      {}
    & {}
    &  \multicolumn{1}{c|}{\B\T $N_1^t$}
    &  \multicolumn{1}{c|}{\B\T $N_1^a$}
    &  \multicolumn{1}{c|}{\B\T $N_1^r$}
    &  \multicolumn{1}{c|}{\B\T $N_1^u$}
    &  \multicolumn{1}{c||}{\B\T $N_{FE}^{\mathrm{net}}$}
    &  \multicolumn{1}{c|}{\B\T $N_2^t$}
    &  \multicolumn{1}{c|}{\B\T $N_2^a$}
    &  \multicolumn{1}{c|}{\B\T $N_2^r$}
    &  \multicolumn{1}{c|}{\B\T $N_2^u$}
    &  \multicolumn{1}{c||}{\B\T $N_{FE}^{\mathrm{net}}$}
    & {}
\\ \hline
{\multirow{4}{*}{{\rule{0pt}{7.0ex}} HAN}}
& \BB \TT A & 575 & 568 & 7 & 132 & 90\,M
& 537 & 492 & 45 & 56 & 60\,M
& 436
\\ \cline{2-13}
{}
& \BB \TT B & 575 & 544 & 31 & 4 & 90\,M
& 706 & 701 & 5 & 161 & 60\,M
& 540
\\ \cline{2-13}
{}
& \BB \TT C & 575 & 543 & 32 & 6 & 90\,M
& 725 & 721 & 4 & 184 & 102.7\,M
& 537
\\ \cline{2-13}
{}
& \BB \TT D & 575 & 538 & 37 & 3 & 90\,M
& 713 & 708 & 5 & 173 & 102.2\,M
& 535
\\ \hline
{\multirow{4}{*}{{\rule{0pt}{7.0ex}} BLA}}
& \BB \TT A & 901 & 884 & 17 & 425 & 90\,M
& 678 & 497 & 181 & 38 & 60\,M
& 459
\\ \cline{2-13}
{}
& \BB \TT B & 901 & 850 & 51 & 4 & 90\,M
& 1034 & 1033 & 1 & 187 & 60\,M
& 846
\\ \cline{2-13}
{}
& \BB \TT C & 901 & 849 & 52 & 0 & 90\,M
& 1036 & 1036 & 0 & 187 & 101.4\,M
& 849
\\ \cline{2-13}
{}
& \BB \TT D & 901 & 849 & 52 & 0 & 90\,M
& 1040 & 1040 & 0 & 191 & 101.2\,M
& 849
\\ \hline
{\multirow{4}{*}{{\rule{0pt}{7.0ex}} NYT}}
& \BB \TT A & 627 & 604 & 23 & 97 & 90\,M
& 573 & 544 & 29 & 37 & 90\,M
& 507
\\ \cline{2-13}
{}
& \BB \TT B & 627 & 591 & 36 & 22 & 90\,M
& 647 & 640 & 7 & 71 & 90\,M
& 569
\\ \cline{2-13}
{}
& \BB \TT C & 627 & 591 & 36 & 28 & 90\,M
& 645 & 633 & 12 & 70 & 113.7\,M
& 563
\\ \cline{2-13}
{}
& \BB \TT D & 627 & 591 & 36 & 22 & 90\,M
& 648 & 640 & 8 & 71 & 113.1\,M
& 569
\\ \hline
{\multirow{4}{*}{{\rule{0pt}{7.0ex}} GOY}}
& \BB \TT A & 489 & 459 & 30 & 123 & 90\,M
& 458 & 401 & 57 & 65 & 90\,M
& 336
\\ \cline{2-13}
{}
& \BB \TT B & 489 & 443 & 46 & 56 & 90\,M
& 545 & 509 & 36 & 122 & 90\,M
& 387
\\ \cline{2-13}
{}
& \BB \TT C & 489 & 447 & 42 & 82 & 90\,M
& 519 & 473 & 46 & 108 & 122.1\,M
& 365
\\ \cline{2-13}
{}
& \BB \TT D & 489 & 443 & 46 & 56 & 90\,M
& 544 & 508 & 36 & 121 & 121.8\,M
& 387
\\ \hline
\end{tabular}
\label{tbl_algo}
\end{table*}

\section{NSGA-II with local search}
\label{sec_nsga2_ls}
In the MOPSO+ scheme mentioned earlier, the current ND
solutions are stored in an archive (usually referred to
as ``external archive" in the literature); local search (LS)
is performed at regular intervals, and new ND solutions
resulting from LS are added to the archive. One
of the solutions in the archive is designated as the global
leader using Roulette-wheel selection, favouring solutions
in the least crowded regions of the archive. The position
of the global leader affects the velocity of particles in
the swarm, and thus the process of local search~-- through
the external archive~-- is coupled with the progress of the
PSO algorithm.

In this work, we explore the effectiveness of local search
when combined with one of the industry-standard MOEAs, viz.,
the NSGA-II algorithm\,\cite{deb2002}, for WDS optimisation.
In the following, we describe how various features can be
added in a step-by-step manner to the NSGA-II algorithm
to finally incorporate local search into the algorithm.
The intermediate algorithms introduced in this process can
also be used as stand-alone algorithms for WDS optimisation.

\begin{list}{(\Alph{cntr2})}{\usecounter{cntr2}}
 \item
  NSGA-II: This is the real-coded
  NSGA-II algorithm\,\cite{deb2002}, modified
  suitably for the WDS problem. The variables take
  on integer values corresponding to the indices
  for pipe diameters, but they are treated as real
  (continuous) variables. In the function evaluation
  step, each of them is converted to the nearest integer,
  following \cite{wang2014}. The algorithm parameters
  $p_c$ and $\eta _c$ are related to crossover, and
  $p_m$ and $\eta _m$ to mutation\,\cite{deb2002}.
  We will denote the population size by $N$, number of
  generations for a specific run by $N_{\mathrm{gen}}$,
  number of independent runs by $N_r$, and the number
  of real parameters (same as the number of pipes in
  the WDS problem) by $N_{\mathrm{real}}$. Note that,
  in each independent run, up to $N$ non-dominated
  solutions are produced by NSGA-II, and the final
  ND set is obtained by combining the ND sets given
  by the $N_r$ independent runs.
 \item
  NSGA-II with external archive: In this scheme\,\cite{patil2018},
  an external archive is used to store ND solutions. The solutions
  stored in the archive do not participate in the evolution of
  the population in any way; the archive is used purely as a
  storage mechanism. A ``fixed hypergrid" without
  boundaries\,\cite{patil2018}, which provides a memory-efficient
  implementation, is used as the external archive. In each
  generation, for each individual of the population not dominated by
  the solutions stored in the archive, a corresponding new solution
  is added to the archive, and any existing solutions in the archive
  which are dominated by this new solution are removed. There is
  no other interaction between the evolving population and
  the external archive. The hypergrid parameters\,\cite{patil2018}
  are selected so that the number of solutions in any hypercell remains
  smaller than the maximum allowed occupancy. This means that
  a current ND solution can get discarded during the evolution process
  only if it gets dominated by an incoming new solution, and not
  because of constraints on the hypergrid. All solutions in the external
  drive are written to a file at the end of a specific run. Note that
  the number of ND solutions in this case~-- even for a single independent run~--
  can be larger than the population size,
  as demonstrated in
  \cite{patil2018}
  for several examples.
 \item
  NSGA-II with external archive and local search: This scheme is similar
  to scheme B except that local search is performed periodically (every
  $N_{LS}$ generations) around each solution stored currently in the
  external archive\,\cite{patil2019}. The archive is updated after the
  LS step by adding new ND solutions arising from LS and removing solutions
  which got dominated by the incoming solutions. Further details about
  implementation of local search for the WDS problem can be found in
  \cite{patil2019}.
 \item
  NSGA-II with external archive, local search, and coupling: In the previous
  scheme, the external archive is (possibly) improved periodically by the
  local search process; however, that improvement does not get coupled to
  the individuals in the evolving population. The purpose of scheme D is to
  provide a way to couple (link) the external archive with the population.
  To this end, we use a mechanism similar to that described in
  \cite{barlow2014}: Every $N_{\mathrm{link}}$ generations, the child
  population is taken from the external archive using Roulette-wheel
  selection (favouring the least crowded regions of the archive) instead
  of using selection, crossover, and mutation. Through this mechanism,
  ND solutions in the archive can influence the evolution of the population.
\end{list}
Although our main interest in this paper is to compare the performance of
algorithms A and D above, it is instructive to also consider algorithms
B anc C for WDS optimisation.

\section{Results and discussion}
\label{sec_results}
We consider four medium-size problems described in
\cite{wang2014},
viz., the HAN, BLA, NYT, and GOY networks. For each of these,
we employ algorithms A-D of
Sec.~\ref{sec_nsga2_ls}.
To compute the network resilience for a given network, we use
the EPANET program\,\cite{rossman2000} as in
\cite{wang2014}.
The NSGA-II algorithm parameter values, taken from
\cite{wang2014}, are
$\eta _c \,$=$\, 15$ (distribution index for crossover),
$\eta _m \,$=$\, 7$ (distribution index for mutation),
$p_c \,$=$\, 0.9$ (crossover rate),
$p_m \,$=$\, 1/N_{\mathrm{real}}$ (mutation rate).
Following
\cite{patil2019},
local search~-- applicable in algorithms C and D~-- is carried out more
frequently in the beginning with $N_{LS} \,$=$\, 100$ from generation
1,000 to 5,000, and with
$N_{LS} \,$=$\, 1,000$ thereafter.
Coupling between the archive and the population~-- applicable in algorithm D~--
is implemented only after the first 1,000 generations.

The selection of the population size $N$, number of independent runs $N_r$,
and number of generations $n_{\mathrm{gen}}$ was made after studying their
effect of the ND set obtained for each network. For example, for the BLA
network, with $N \,$=$\, 200$ and $n_{\mathrm{gen}} \,$=$\, 15,000$, it was
observed that increasing $N_r$ beyond 20 did not produce any improvement in
the ND set, and it was therefore fixed at 20. The following parameter
values were selected: (a)~$N \,$=$\, 200$ for all networks,
(b)~$n_{\mathrm{gen}} \,$=$\, 10,000$ for the HAN network and 15,000 for
the other three,
(c)~$N_r \,$=$\, 20$ for the BLA network and 30 for the others.
It should be mentioned that, although a more systematic selection of the
above parameters is desirable, it is not expected to alter the conclusions
of the present study significantly.

To assess the performance of any of the algorithms (A-D) of
Sec.~\ref{sec_nsga2_ls}, we compare the ND set produced by that algorithm
with the benchmark UExeter ND set\,\cite{wang2014}.
First, we present the effect of the parameter
$N_{\mathrm{link}}$
of algorithm D in Table~\ref{tbl_nlink}. This parameter determines the
frequency of interaction between the evolving population and the
archive. In the extreme case of
$N_{\mathrm{link}} \,$=$\, 1$,
the child population in every generation is taken from the archive.
From the table, we see that
$N_{\mathrm{link}} \,$=$\, 1$
generally gives poor results. For example, consider the HAN network.
With
$N_{\mathrm{link}} \,$=$\, 1$,
44 of the benchmark solutions (the $N_1^u$ column) have not been
covered by algorithm D whereas With
$N_{\mathrm{link}} \,$=$\, 10$,
that number drops to 4. We notice also that, for the HAN network, increasing
$N_{\mathrm{link}}$ results in a larger number of unique solutions ($N_2^u$).
However, in general, we see that
$N_{\mathrm{link}} \,$=$\, 10$, 50, and 100 give similar results. In the following,
we use a fixed value
$N_{\mathrm{link}} \,$=$\, 100$.

\begin{figure*}[!ht]
\centering
\scalebox{0.9}{\includegraphics{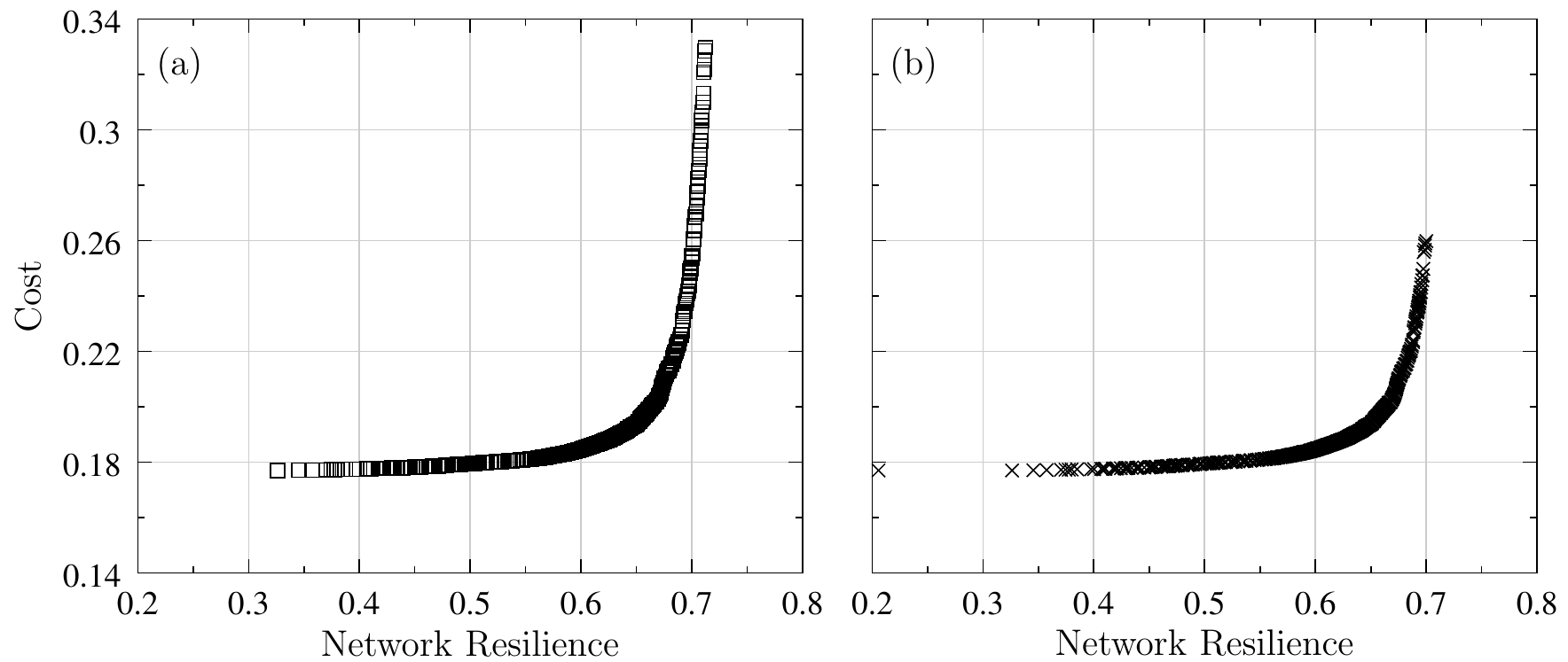}}
\caption{Non-dominated solutions obtained for the GOY network:
(a)~UExeter\,\cite{wang2014}, (b)~Algorithm B.}
\label{fig_goy}
\end{figure*}

The results obtained with algorithms A-D of
Sec.~\ref{sec_nsga2_ls} are summarised in Table~\ref{tbl_algo}.
We can make the following observations from this table.
\begin{list}{(\alph{cntr2})}{\usecounter{cntr2}}
 \item
  Very significant improvement is obtained by algorithm B
  over algorithm A (NSGA-II) for the same computational effort
  $N_{FE}^{\mathrm{net}}$. This means that simply storing all
  ND positions visited by the population is greatly beneficial.
  For example, for the BLA network, NSGA-II could not cover 425 of
  the UExeter solutions whereas algorithm B missed only 4 of the UExeter
  solutions.
 \item
  For the HAN network, the use of local search (algorithm C) gave 184
  unique solutions (not found in the UExeter set) whereas algorithm B
  gave 161, thus pointing to the effectiveness of local search for this
  problem. However, for other problems, local search either did not
  improve the ND set (over algorithm B) or made it worse.
 \item
  For the BLA, NYT, and GOY networks, local search together with coupling
  the population and archive (algorithm D) has produced a larger number of
  unique solutions as compared to only local search (algorithm C).
 \item
  The most significant improvement in the ND set comes from the use of
  external archive (compare the algorithm A and B results).
 \item
  For the GOY network, NSGA-II (algorithm A) as well as the proposed
  modifications of NSGA-II (algorithms B, C, D) are unable to cover a substantial
  number of UExeter solutions. Fig.~\ref{fig_goy} compares the UExeter
  ND set with that obtained with algorithm B. Note that a large number
  of UExeter solutions in the high resilience (or high cost) region are
  missed out by algorithm B (as also by algorithms C and D). 
  As mentioned in \cite{wang2014}, NSGA-II generally captured solutions in
  the low- and medium-cost regions but not in the high-cost regions.
  With algorithms B, C, D, this drawback could be eliminated for the
  HAN and BLA networks and to some extent for the NYT network. However,
  for the GOY network, none of the modifications is effective in obtaining
  the high-cost region of the PF.
\end{list}

\section{Conclusions}
\label{sec_conclusions}
In conclusion, three step-by-step modifications of the NSGA-II
algorithm have been presented in this work. The new algorithms
have been used for the medium-size water network problem described in
\cite{wang2014}.
For three of the four problems, the proposed algorithms have given
substantial improvement over the best-known Pareto fronts (ND sets)
available in the literature. It was found that the most significant
contribution in this improvement arises from the use of an external
archive to store all ND positions visited by the population.

Compared to the recently proposed MOPSO+ algorithm\,\cite{patil2019},
the algorithms presented in this work are found to be less effective
for the two-objective WSD optimisation problem of
\cite{wang2014} (see
Tables~\ref{tbl_mopso_mbp} and
\ref{tbl_algo}).
A mechanism other than that described in this paper for coupling the
archive and the evolving population needs to be explored for improved
performance.

\bibliographystyle{IEEEtran}
\bibliography{arx2}

\vfill

\end{document}